# Comparing Robustness of Pairwise and Multiclass Neural-Network Systems for Face Recognition

J. Uglov, V. Schetinin, C. Maple

Computing and Information System Department, University of Bedfordshire, Luton, UK

**Abstract.** Noise, corruptions and variations in face images can seriously hurt the performance of face recognition systems. To make such systems robust, multiclass neural-network classifiers capable of learning from noisy data have been suggested. However on large face data sets such systems cannot provide the robustness at a high level. In this paper we explore a pairwise neural-network system as an alternative approach to improving the robustness of face recognition. In our experiments this approach is shown to outperform the multiclass neural-network system in terms of the predictive accuracy on the face images corrupted by noise.

## 1. Introduction

Performance of face recognition systems is achieved at a high level when such systems are robust to noise, corruptions and variations in face images [1]. To make face recognition systems robust, multiclass artificial neural networks (ANNs) capable of learning from noisy data have been suggested [1]. However on large face data sets such neural-network systems cannot provide the robustness at a high level [1] - [3]. To overcome this problem pairwise classification systems have been proposed, see e.g. [3], [4].

   In this paper we explore a pairwise neural-network system as an alternative approach to improving the robustness. In our experiments this approach is shown to outperform the multiclass neural-network system in terms of the predictive accuracy on the face image data described in [5].

   In section 2 we briefly describe face image representation and noise problems, and then in section 3 we describe a pairwise neural-network system proposed for face recognition. Section 4 describes our experiments and finally section 5 concludes the paper.

## 2. Face Image Representation and Noise Problems

Following to [1] – [3], we use the principal component analysis (PCA) to represent face images as $m$-dimensional vectors of components. The PCA is the common technique for data representation in face recognition systems.

   The first two principal components which make the most important contribution in face recognition can be used to visualise the scatter of patterns of different classes (faces). Therefore the use of such a visualisation allows us to observe how the noise can corrupt the boundaries of classes. For example, Fig. 1 shows two graphs depicting the examples of four classes whose centres of gravity are visually distinct. The left side plot depicts the examples taken from the original data while the right side plot depicts these examples containing noise components drawn from a Gaussian density function with zero mean and the standard deviation alpha = 0.5.

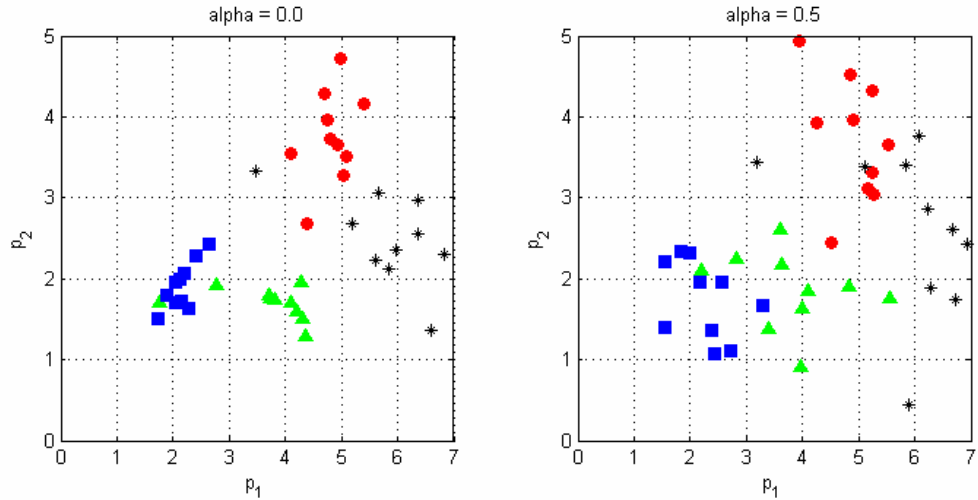

Fig. 1. An example of scattering the samples drawn from the four classes for alpha = 0 (the left side) and alpha = 0.5 (the right side) in a plane of the first two principal components $p_1$ and $p_2$.

From this plot we can observe that the noise components corrupt the boundary of the given classes, and therefore the performance of a face recognition system can be affected. From these plots we can also observe that the boundaries between pairs of the classes can remain almost the same. This inspire us to exploit such a classification scheme to implement a pairwise neural-network system for face recognition.

## 3. A Pairwise Neural-Network System

The idea behind the pairwise classification is to use two-class ANNs learning to classify all possible pairs of classes. Therefore for $C$ classes the pairwise system should include $C*(C-1)/2$ ANNs learnt to solve two-class problems.

For example, for classes $\Omega_1$, $\Omega_2$, and $\Omega_3$ depicted in Fig. 2, the number of two-class ANNs is equal to 3. In this figure the lines $f_{1/2}$, $f_{1/3}$ and $f_{2/3}$ are the dividing hyperplanes learnt by the ANNs. We can simply assume these functions are given the positive values for examples of the classes standing first in the lower indexes (1, 1, and 2) and the negative values for the classes standing second in there (2, 3, and 3).

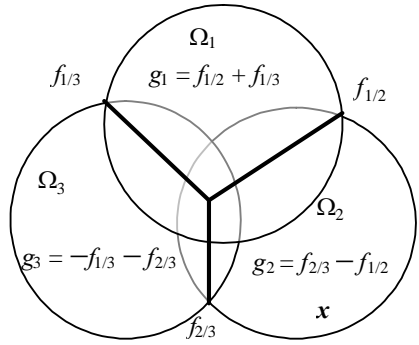

Fig. 2. Hyperplanes $f_{1/2}$, $f_{1/3}$ and $f_{2/3}$ learnt to divide the following pairs of classes: $\Omega_1$ versus $\Omega_2$, $\Omega_1$ versus $\Omega_3$, and $\Omega_2$ versus $\Omega_3$.

Now we can combine hyperplanes $f_{1/2}$, $f_{1/3}$ and $f_{2/3}$ to build up the new dividing hyperplanes $g_1$, $g_2$, and $g_3$. The first hyperplane $g_1$ combines the functions $f_{1/2}$ and $f_{1/3}$ so that $g_1 = f_{1/2} + f_{1/3}$. These functions are taken with weights of 1.0 because both functions $f_{1/2}$ and $f_{1/3}$ give the positive output values on the examples of class $\Omega_1$. Likewise, the second and third hyperplanes are as follows: $g_2 = f_{2/3} - f_{1/2}$ and $g_3 = -f_{1/3} - f_{2/3}$.

In practice each of hyperplanes $g_1$, ..., $g_C$, can be implemented as a two-layer feed-forward ANN with a given number of hidden neurons fully connected to the input nodes. Then we can introduce the output neuron summing all outputs of the ANNs to make a final decision.

For example, the pairwise neural-network system depicted in Fig 3 consists of three neural networks performing the functions $f_{1/2}$, $f_{1/3}$, and $f_{2/3}$. The three output neurons $g_1$, $g_2$, and $g_3$ are connected to these networks with weights equal to (+1, +1), (−1, +1) and (−1, −1), respectively.

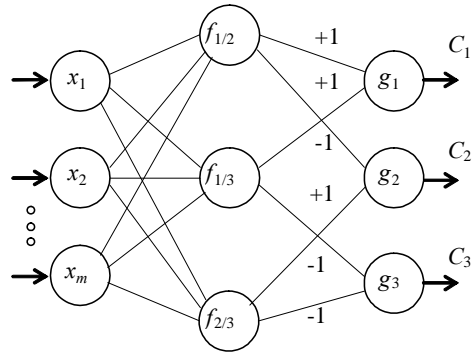

Fig. 3. An example of pairwise neural-network system for $C = 3$ classes.

In general, a pairwise neural-network system consists of $C(C-1)/2$ neural networks, performing functions $f_{1/2}$, ..., $f_{i/j}$, ..., $f_{C-1/C}$, and $C$ output neurons $g_1$, ..., $g_C$, where $i < j = 2, ..., C$. We can see that the weights of output neuron $g_i$ connected to the hidden neurons $f_{i/k}$ and $f_{k/i}$ should be equal to $+1$ and $-1$, respectively.

## 4. Experiments

The goal of our experiments is to compare the robustness of the proposed pairwise and standard multiclass neural-network systems on the Cambridge ORL face image data set [5] (in a full paper, the experiments will run on different face image data sets). To estimate the robustness we add noise components to the data and then estimate the performance on the test data within 5 fold cross-validation. The performances of the pairwise and multiclass systems are listed in Table 1 and shown in Fig. 4.

Table 1. Performance of the pairwise (P) and multiclass (M) neural-network systems over alpha. The performances are represented by the means and $2\sigma$ intervals.

| alpha | 0.0 | 0.1 | 0.3 | 0.5 | 0.7 | 0.9 | 1.1 | 1.3 |
|---|---|---|---|---|---|---|---|---|
| P, mean | 0.972 | 0.966 | 0.953 | 0.920 | 0.859 | 0.772 | 0.659 | 0.556 |
| P, $2\sigma$ | ±0.004 | ±0.013 | ±0.017 | ±0.013 | ±0.018 | ±0.030 | ±0.028 | ±0.031 |
| M, mean | 0.952 | 0.951 | 0.932 | 0.898 | 0.802 | 0.678 | 0.557 | 0.419 |
| M, $2\sigma$ | ±0.017 | ±0.016 | ±0.025 | ±0.016 | ±0.015 | ±0.052 | ±0.036 | ±0.050 |

From this table we can see that for alpha ranging between 0.0 and 1.3 the proposed pairwise system significantly outperforms the multiclass systems. For alpha = 0.0 the improvement in the performance is 2.0% while for alpha = 1.1 the improvement becomes 10.2%.

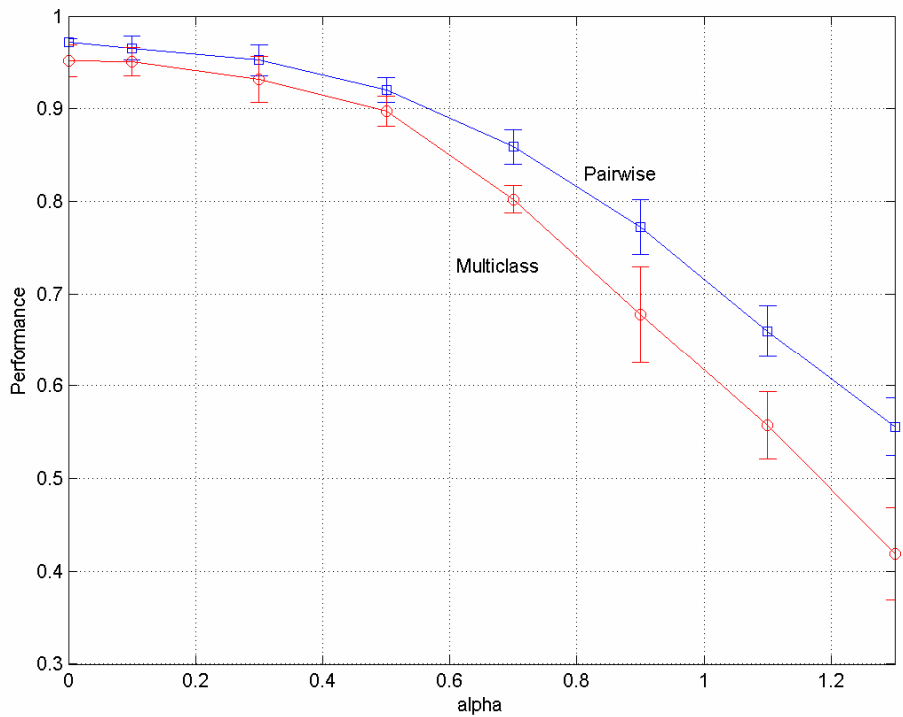

Fig. 4. Performance of the pairwise and multiclass neural-network systems over alpha. Solid lines and bars are the mean and $2\sigma$ intervals, respectively.

## 5. Conclusion

We have proposed a pairwise neural-network system for face recognition in order to reduce the negative effect of noise and corruptions in face images. Within such a classification scheme we expect that the improvement in the performance can be achieved on the base of our observation that boundaries between pairs of classes remain almost the same while a noise level increases.

We have compared the performances of the proposed pairwise and standard multiclass neural-network systems on the face dataset [5]. Evaluating the mean values and standard deviations of the performances under different levels of noise in the data, we have found that the proposed pairwise system is superior to the multiclass neural-network system.

Thus we conclude that the proposed pairwise system is capable of decreasing the negative effect of noise and corruptions in face images. Clearly this is a very desirable property for face recognition systems when the robustness is of crucial importance.